\documentclass[letterpaper]{article} 
\usepackage{aaai2026}  
\usepackage{times}  
\usepackage{helvet}  
\usepackage{courier}  
\usepackage[hyphens]{url}  
\usepackage{graphicx} 
\usepackage{natbib}  
\usepackage{caption} 
\usepackage{algorithm}
\usepackage{algorithmic}

\usepackage{amsmath,amssymb,amsfonts} 
\usepackage{multirow}
\usepackage{bm}
\usepackage{subcaption}
\usepackage{booktabs}
\usepackage{tikz}
\usetikzlibrary{calc,patterns,angles,quotes}
\usetikzlibrary{shapes.geometric, arrows}
\usetikzlibrary{fit,backgrounds} 
\usetikzlibrary{spy,calc}

\usepackage{pgfplots}
\usepackage{pgfplotstable}
\pgfplotsset{compat=newest}
\usepackage{tikz-3dplot}
\usepackage{arydshln}
\usepackage{newfloat}
\usepackage{listings}
\DeclareCaptionStyle{ruled}{labelfont=normalfont,labelsep=colon,strut=off} 
\floatstyle{ruled}
\newfloat{listing}{tb}{lst}{}
\floatname{listing}{Listing}
\title{RiemanLine: Riemannian Manifold Representation of 3D Lines for Factor Graph Optimization}
\author{
      Yan Li\textsuperscript{\rm 1}, 
      Ze Yang\textsuperscript{\rm 2}, 
      Keisuke Tateno\textsuperscript{\rm 3}, 
      Federico Tombari\textsuperscript{\rm 3,\rm 4}, 
      Liang Zhao\textsuperscript{\rm 5}, 
      Gim Hee Lee\textsuperscript{\rm 1} 
}
\affiliations{
    \textsuperscript{\rm 1}National University of Singapore\\
    \textsuperscript{\rm 2}Peking University\\
    \textsuperscript{\rm 3}Google\\
    \textsuperscript{\rm 4}Technical University of Munich\\
    \textsuperscript{\rm 5}The University of Edinburgh\\

}

\usepackage{bibentry}

\begin{document}

\maketitle

\begin{abstract}
Minimal parametrization of 3D lines plays a critical role in camera localization and structural mapping. 
Existing representations in robotics and computer vision predominantly handle independent lines, 
overlooking structural regularities such as sets of parallel lines that are pervasive in man-made environments. 
This paper introduces \textbf{RiemanLine}, a unified minimal representation for 3D lines formulated on Riemannian manifolds that jointly accommodates both individual lines and parallel-line groups. 
Our key idea is to decouple each line landmark into global and local components: 
a shared vanishing direction optimized on the unit sphere $\mathcal{S}^2$, 
and scaled normal vectors constrained on orthogonal subspaces, enabling compact encoding of structural regularities. 
For $n$ parallel lines, the proposed representation reduces the parameter space from $4n$ (orthonormal form) to $2n+2$, naturally embedding parallelism without explicit constraints. 
We further integrate this parameterization into a factor graph framework, allowing global direction alignment and local reprojection optimization within a unified manifold-based bundle adjustment. 
Extensive experiments on ICL-NUIM, TartanAir, and synthetic benchmarks demonstrate that our method achieves significantly more accurate pose estimation and line reconstruction, while reducing parameter dimensionality and improving convergence stability. 
\end{abstract}

\begin{links}
    \link{Code}{https://github.com/yanyan-li/RiemanLine}
\end{links}

\section{Introduction}

\begin{figure}[!t]
\includegraphics[width=\linewidth]{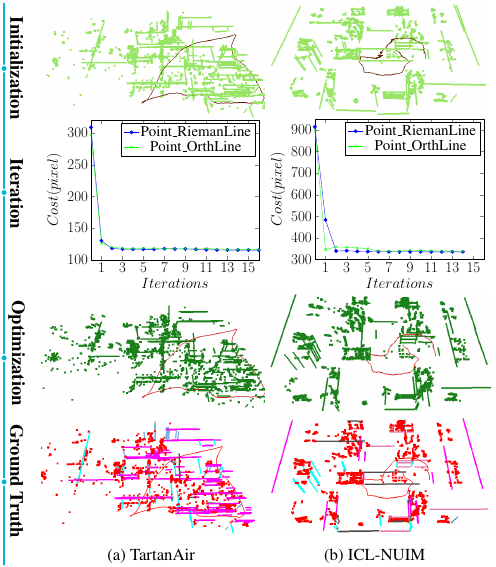}
    \caption{An illustration of co-visibility factor graph optimization based points and lines. The initial factor graphs are depicted in the first row, with landmarks and trajectories colored light green and scarlet, respectively. The convergence curves for different representations are plotted in the second row. The optimized results based on the proposed \textit{Point\_RiemanLine} method and ground-truth graphs are presented in the last rows, respectively.}
    \label{fig:teaser}
\end{figure}

Robustly reconstructing~\cite{schonberger2016structure,forster2014svo} unknown three-dimensional scenes and estimating~\cite{carlone2015initialization,carlone2018convex} six degrees-of-freedom (6-DoF) camera poses from visual inputs are fundamental challenges in robotics and computer vision.
However, these odometry and SLAM methods often suffer from structural inaccuracies and pose drift during the incremental camera tracking and mapping process. To mitigate these issues, techniques such as local bundle adjustment~\cite{mur2015orb,rosinol2020kimera}, sliding window optimization~\cite{qin2018vins,engel2017direct}, and loop closure~\cite{labbe2019rtab,mur2015orb} are commonly incorporated. 
The core of these techniques lies in the use of factor graph optimization~\cite{carlone2018convex}, which jointly refines the structure and transformation parameters. 
In this paper, we present a novel solution to the line-leveraged factor graph optimization problem by proposing a compact line parameterization on a Riemannian manifold along with constraint factors that connect line landmarks to camera poses, as shown in Figure~\ref{fig:teaser}.

Point features have long served as the foundation for most visual pose estimation systems, as demonstrated by their widespread adoption in leading methods~\cite{mur2015orb,qin2018vins,rosinol2020kimera}. Despite their success, point features exhibit notable limitations, especially in challenging environments such as indoor scenes. To overcome the limitations of point-only factor graph optimization~\cite{li2023open}, recent efforts have explored the integration of additional geometric primitives such as lines~\cite{lu2015visual,zuo2017robust} and planes~\cite{zhou2021lidar} into both tracking and optimization modules. Compared to plane detection which is typically based on depth maps~\cite{salas2014dense} or convolutional neural networks~\cite{paigwar2020gndnet}, line features can be efficiently extracted from RGB images, offering a versatile and computationally efficient means to enhance visual odometry performance. 
The most popular and widely adopted line parameterization method in line-based SLAM systems is the \textbf{Orthonormal algorithm}~\cite{bartoli2005structure}, which enables elegant optimization within the framework of \textbf{Lie Algebra}.

Typically, a single line segment contributes a re-projection factor~\cite{hartley2003multiple} to the optimization module, whereas a collection of lines imparts broader structural and global regularities. Specifically, a group of parallel line segments on a 2D image plane crosses at the vanishing point, which can be projected to the camera coordinate to obtain the corresponding vanishing direction~\cite{mclean1995vanishing} via a related intrinsic matrix. 
A group of vanishing direction vectors can model structured environments such as the Atlanta or Manhattan World assumptions~\cite{straub2017manhattan}. However, these assumptions are often too restrictive for general environments, and it is also difficult to optimize the structure as one primitive in factor graph optimization modules. 
Furthermore, conventional minimal parameterization~\cite{bartoli2005structure} that is widely adopted in line-SLAM systems~\cite{zuo2017robust,he2018pl} lacks the expressiveness to compactly parametrize a group of structure lines. 
Consequently, this leads to increased complexity and reduced efficiency since the optimization frameworks must rely on additional parameters and manually introduce constraints into loss functions to capture the relationships among these lines.

In this paper, we first propose a unified minimal representation for 3D lines, including individual and structure line landmarks, which has a clear geometric interpretation in representation and optimization based on Riemannian manifolds. 
Specifically, the line parameters are explicitly decoupled into \textbf{global} and \textbf{local} components: the global component defines the shared direction of parallel lines, and the local component that lies in the orthogonal plane encodes scaled normals that distinguish each line.
Representing $n$ parallel 3D lines with the Orthonormal representation requires $4n$ parameters. In contrast, our method reduces this to only $2n + 2$ parameters, significantly reducing the dimensionality while implicitly encoding parallelism without additional constraints. We integrate this representation into a joint factor graph framework with co-visibility factors, enabling both accurate pose estimation and structurally consistent line reconstruction.
Our contributions are summarized as follows:
\begin{itemize}
    \item We introduce a unified minimal parameterization for 3D lines on Riemannian manifolds that seamlessly extends from independent lines to sets of parallel lines by explicitly decoupling global and local components.
    \item A joint factor graph framework incorporating co-visibility and extensibility factors, specifically designed to leverage the proposed minimal representation;
    \item We conduct extensive evaluations on ICL-NUIM, TartanAir, and synthetic datasets, demonstrating that our method achieves higher accuracy with fewer parameters and improved convergence stability compared to existing representations.
\end{itemize}

\section{Related work}

\begin{table}
    \centering
    \resizebox{0.9\linewidth}{!}{
    \begin{tabular}{l|c}
     \toprule
        Algorithm & Parameters \\ \hline
        Euclidean & \begin{tabular}[c]{@{}c@{}}$\mathbf{P}_s$, $\mathbf{P}_e$ \\ endpoints of the 3D line\end{tabular} \\ \hline
        Pl\"ucker & \begin{tabular}[c]{@{}c@{}}$\mathbf{n}$, $\mathbf{v}$ \\ $\mathbf{n}=[\mathbf{P}_s]_{\times}\mathbf{P}_e$, $\mathbf{v}=\mathbf{P}_s-\mathbf{P}_e$  \end{tabular}  \\ \hline
        Quaternion &\begin{tabular}[c]{@{}c@{}} $\bar{\mathbf{q}}$, $d$ \\ $\mathcal{R}(\bar{\mathbf{q}}) = [\mathbf{n} \; \mathbf{v} \;[\mathbf{n}]_{\times}\mathbf{v}]$, $d = ||\mathbf{n}||/||\mathbf{v}||$ \end{tabular} \\ \hline
        Closest Point &$\mathbf{j} = d\bar{\mathbf{q}}$   \\ \hline
        Spherical & \begin{tabular}[c]{@{}c@{}}$\theta$, $\phi$, $\alpha$, $d$ \\ $\theta$, $\phi$, and $\alpha$ are angles, d is the distance   \end{tabular}  \\
     \bottomrule
    \end{tabular}}
    \caption{Popular line representations and transformation relationships between them. Here $[\cdot]_{\times}$ is the skew-s operation, and the $\mathcal{R}$ is the transformation operation from quaternion to rotation matrix. The definitions of the remaining parameters are provided in Section~\ref{sec:line_represent}.}
    \label{tab:line_represents}
\end{table}

As shown in Table~\ref{tab:line_represents}, two stages of representation are typically involved in landmark reconstruction and optimization tasks~\cite{bartoli2005structure}. The first stage focuses on 3D triangulation from 2D measurements, and the second seeks a minimal parametrization for iterative refinement. The two stages can be unified into a single step when the degrees of freedom used in reconstruction meet the requirements of minimal parametrization.

\begin{figure*}
\centering
\begin{subfigure}{0.45\linewidth}
   \includegraphics[width=\linewidth]{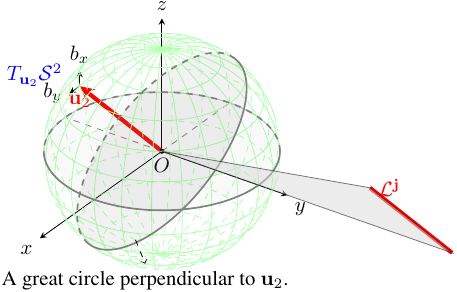}
   \caption{$\mathcal{S}^2$ Space}
   \label{subfig:rieman_s2}
\end{subfigure}
\begin{subfigure}{0.45\linewidth}
  \includegraphics[width=\linewidth]{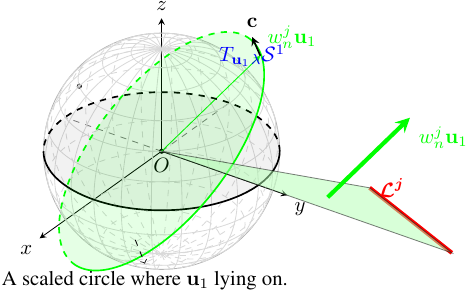}
   \caption{${_\lambda}\mathcal{S}^1$ Space}
   \label{subfig:line_s1}
\end{subfigure}
\caption{Illustration of the proposed parametrization for a line landmark $\mathbf{\mathcal{L}}^j$. The vanishing direction vector $\mathbf{u}_2$ ($\|\mathbf{u}_2\|=1$) and scaled normal vector $\omega_{n}^j\mathbf{u}_1$ ( $\|\mathbf{u}_1\|=1, \omega_{n}^j>0$) are optimized on the tangent spaces $T_{\mathbf{u}_2}\mathcal{S}^2$ of the sphere and $T_{\mathbf{u}_1}{_\lambda}\mathcal{S}^{1}$ of the scaled circle, respectively. }
\label{fig:single_line}
\end{figure*} 

Generally, Euclidean \textit{XYZ}~\cite{mur2015orb,qin2018vins} is used to parametrize endpoints of a finite line in the 3D space. Although this representation can be used to provide re-projection residuals of lines for camera pose optimization, it has the over-parameterization problems in landmark optimization. 
Similar to the \textit{Euclidean}, a line at infinity can be represented as the \textit{Pl\"ucker} coordinate via two three-dimensional vectors containing the direction of the line and normal based on the line and the camera coordinate frame as listed in Table~\ref{tab:line_represents}. 
Furthermore, the dual quaternion approach~\cite{kottas2013efficient} represents a line in 3D space using dual quaternions, which provides a concise way to represent rotations and translations in 3D space and can be used to encode the position and orientation of lines. The advantage of this approach is that it allows for easy composition of transformations, which makes it useful in applications such as robotics and animation. 
By multiplying the unit quaternion and the distance parameter, the \textit{closest point} method~\cite{8793507} can be considered as the 'closest point' for a 3D line. The transformation relationships between \textit{Pl\"ucker}, \textit{Quaternion} and \textit{Closest Point} methods are listed in Table~\ref{tab:line_represents}. The rotation matrix used in those representations can be optimized via \textit{Lie algebras} which define the tangent spaces of related manifolds. 
For the \textit{Spherical} form, a line can also be represented by three angles and a distance parameter. 
Instead of optimizing the orthogonal matrix through \textit{Lie Groups} being special instances of the manifold, a more general Riemannian manifold is used to refine the vanishing and normal vectors subsequently in the proposed method.

Structure-SLAM~\cite{li2020structure} estimates the orientation using surface normals from monocular images, while Linear-SLAM~\cite{joo2021linear} extracts planes from depth maps. Given a known rotation, the translation estimation becomes linear. The Manhattan World (MW) model assumes orthonormal landmarks and the Atlanta World (AW) model introduces multiple perpendicular horizontal directions. 
The multi-MW model~\cite{yunus2021manhattanslam} enforces orthogonality within local regions instead of throughout the scene. However, such methods often neglect optimization within factor graphs.
Recent approaches integrate structural constraints, lines and planes, directly into optimization. \cite{zhang2015building} pioneered using parametric 3D lines as SLAM landmarks, enhancing traditional point-based methods. \cite{lu2015visual} extended this by incorporating diverse structural features within a multilayer feature graph (MFG) to improve environmental representation and pose estimation. 
Kimera-VIO~\cite{rosinol2020kimera} leverages 3D meshes to enforce coplanarity constraints within factor graphs. PLP-VIO~\cite{li2020leveraging} further refines this approach by incorporating line-based meshes to enhance spatial understanding.
CoP~\cite{li2020co} introduced a novel parameterization that represents points and lines using plane parameters to preserve geometric consistency during optimization.
Beyond scene-specific models, Struct-VIO~\cite{zhou2015structslam, zou2019structvio} mitigates directional errors and reducing drift by parameterizing line segments parallel to the local Manhattan world.
Our method generalizes this by parameterizing all parallel lines, which uses the structural information to improve pose estimation accuracy in SLAM.

\section{On-Manifold Representation of 3D Lines}\label{sec:line_represent}

\subsection{Preliminary: Representation in \textit{Orthonormal}}

As shown in Table~\ref{tab:line_represents}, a 3-dimensional finite line \(\mathbf{L}^w\) in the world coordinates can be represented by its two 3D endpoints, \(\mathbf{P}^w_s\) and \(\mathbf{P}^w_e\), as \(\mathbf{L}^w = \left[\begin{array}{cc} \mathbf{P}^w_s & \mathbf{P}^w_e \end{array}\right]\). For an infinite line, the direction and normal vectors can be used to represent the line \(\bm{\mathcal{L}^w} = \left[\begin{array}{cc} \mathbf{n}^w & \mathbf{d}^w \end{array}\right]\) based on the $\textit{Pl\"ucker}$ representation, where \(\mathbf{n}^w\) and \(\mathbf{d}^w\) can be derived from the endpoints \(\mathbf{P}^w_s\) and \(\mathbf{P}^w_e\). This can be further expressed as the multiplication of several matrices via the following function:
\begin{equation}
\begin{split}
     \bm{\mathcal{L}}^w &= \left[\begin{array}{cc} \mathbf{n}^w & \mathbf{d}^w \end{array}\right] \\ 
     &=   \left[ \begin{array}{cc} \frac{\mathbf{n}^w}{||\mathbf{n}^w||} & \frac{\mathbf{d}^w}{||\mathbf{d}^w||}\end{array} \right] \left[ \begin{array}{cc}
     ||\mathbf{n}^w||&0  \\
     0& ||\mathbf{d}^w||
\end{array} \right] \\
\end{split}.
\end{equation}
%
We then use  
$\left[ \begin{array}{cc} \frac{\mathbf{n}^w}{||\mathbf{n}^w||} & \frac{\mathbf{d}^w}{||\mathbf{d}^w||}\end{array} \right]$ and $\left[ \begin{array}{cc}
     ||\mathbf{n}^w||&0  \\
     0& ||\mathbf{d}^w||\end{array} \right]$ to establish the following matrices:
 \begin{equation}
     \left\{\begin{split}
         \mathbf{U} &= \left[\begin{array}{ccc}
     \frac{\mathbf{n}^w}{||\mathbf{n}^w||}& \frac{\mathbf{d}^w}{||\mathbf{d}^w||} &\frac{\mathbf{n}^w \times{\mathbf{d}^w } }{||\mathbf{n}^w\times{\mathbf{d}^w }||} \\
\end{array} \right]; \\ 
    \mathbf{W} &= \left[ \begin{array}{cc}
     ||\mathbf{n}^w||/\lambda&-||\mathbf{d}^w||/\lambda  \\
     ||\mathbf{d}^w||/\lambda& ||\mathbf{n}^w||/\lambda
\end{array} \right]. \\ 
     \end{split} \right.
 \end{equation}
Here, $\lambda$ denotes $\sqrt{||\mathbf{d}^w||^2+||\mathbf{n}^w||^2}$, $\mathbf{U}\in SO(3)$, and $\mathbf{W}\in SO(2)$.

In \textit{quaternion-based} optimization methods~\cite{8793507}, $\mathbf{U}$ is represented as a quaternion vector via $\bar{\mathbf{q}} = \mathcal{R}^{-1}(\mathbf{U})$, and the optimization process is implemented based on quaternion representation. With the widespread application of differentiable manifolds in SLAM optimization problems, the Orthonormal representation~\cite{bartoli2005structure} maps the update process of $SO(3)\times{SO(2)}$ on the tangent space $\textit{Lie algebra}$ based on 4 degree-of-freedom $\left[\begin{array}{cc}
     \bm{\rho}^T& \omega
\end{array} \right]^T$ that is the minimal line representation widely used in point-line SLAM~\cite{zuo2017robust,li2020co} and VIO~\cite{he2018pl} systems.

\subsection{Riemannian Representation of a Single 3D Line}

We decouple the single line $\bm{\mathcal{L}^j}$ into two components: the unit direction vector $\mathbf{u}_2$, and the scaled normal vector $\omega_n^j\mathbf{u}_1$. 
As illustrated in Figure~\ref{fig:single_line}, the direction vector $\mathbf{u}_2$ (with $\|\mathbf{u}_2\| = 1$) lies on the surface of the unit sphere $\mathcal{S}^2$. The scaled normal vector $\omega_n^j\mathbf{u}_1$ lies on the plane $_\lambda\mathcal{S}^1$, which is centered at the origin and orthogonal to $\mathbf{u}_2$.

\textbf{Optimization on $\mathcal{S}^2$.}
As illustrated in Figure~\ref{subfig:rieman_s2}, the tangent space to $\mathcal{S}^2$ (at $\mathbf{u}_2$) is given by: 
\begin{equation}
    T_{\mathbf{u}_2}\mathcal{S}^2 \dot{=} \{ \mathbf{x}\in\mathbb{R}^3| \mathbf{x}^T\mathbf{u}_2 = 0   \}. 
\end{equation}
Here, $\mathbf{x}^T\mathbf{u}_2 = 0$ shows that the vectors $\mathbf{x}$ in the tangent space are perpendicular to $\mathbf{u}_2$. Although $\mathcal{S}^2$ is a non-linear 3D space, its tangent space $T_{\mathbf{u}_2}\mathcal{S}^2$ is a 2-DoF linear space. Based on the perpendicular bases $\mathbf{b}_x$ and $\mathbf{b}_y$, any vector adjacent to $\mathbf{u}_2$ on the unit sphere can be represented.

In the representation process, $\delta\theta_1$ and $\delta\theta_2$, $\left[ \begin{array}{cc}
     \delta\theta_1& \delta\theta_2
\end{array}\right]^T=\delta\bm{\theta}$, are corresponding disturbances towards $\mathbf{b}_x$ and $\mathbf{b}_y$, respectively. Consequently, the perturbation $\delta\mathbf{m}$ on the tangent space can be represented as:
\begin{equation}
    \delta\mathbf{m} = \left[\begin{array}{cc}
         \mathbf{b}_x& \mathbf{b}_y
    \end{array}\right] \left[\begin{array}{c}
        \delta\theta_1 \\
        \delta\theta_2
    \end{array} \right],
    \label{eq:delta_m}
\end{equation}
and the additive operation on the unit sphere is denoted as $\mathbf{u}_2 \boxplus \delta\mathbf{m}$, which maps $\delta\mathbf{m}$ back to $\mathbf{S}^2$ via the $\textit{Riemannian exponential}$ $Exp_{\mathbf{u}_2}(\delta\mathbf{m})$, and the vector after perturbation is denoted as $\mathbf{v}_2$ computed via:
\begin{equation}
\begin{split}
     \mathbf{v}_2 &= \mathbf{u}_2 \boxplus \delta\mathbf{m}  \\ &=\mathbf{u}_2\cos{||\delta\mathbf{m}||} + \frac{\delta\mathbf{m}}{\|\delta\mathbf{m}\|}\sin{\|\delta\mathbf{m}\|}, 
\end{split}
\label{eq:update_v2}
\end{equation}
where the perturbation $\delta{\mathbf{m}}$ is close to $\mathbf{0}$.

\textbf{Optimization on $_{\lambda}\mathcal{S}^1$.}
Concurrently, the tangent space to $_{\lambda}\mathcal{S}^1$ is denoted as: 
\begin{equation}
    T_{\mathbf{u}_1} {_\lambda}\mathcal{S}^1 \dot{=} \{ \mathbf{x}\in \mathbb{R}^3 \;|\; \mathbf{x}^T\mathbf{u}_1 = \mathbf{x}^T\mathbf{u}_2 = 0\}.
\end{equation}
Here, $\mathbf{x}^T\mathbf{u}_1 = \mathbf{x}^T\mathbf{u}_2 = 0$ shows that the vectors on the tangent space are perpendicular to $\mathbf{u}_1$ and $\mathbf{u}_2$ at the same time. 
Although ${_\lambda}\mathcal{S}^1$ is a non-linear 2D space, there is a base vector $\mathbf{u}_3$ which is orthogonal to both $\mathbf{u}_1$ and $\mathbf{u}_2$. 
This base vector can be selected as $\mathbf{u}_3$ 
since $\mathbf{u}_3= \mathbf{u}_2\times{\mathbf{u}_1}$.  
With the perturbation of $\mathbf{u}_2$, the tangent space of $\mathbf{u}_1$ is also perturbed by the orthogonal relationship.
Based on the updated global direction $\mathbf{v}_2$, we define the space $T_{\mathbf{u}_1} {_\lambda}\mathcal{S}^1$ based on $\mathbf{u}_3  = \mathbf{v}_2  \times \mathbf{u}_1$.

For a small angle $\delta\gamma$,  we rotate $\mathbf{u}_1$ within the plane orthogonal to $\mathbf{v}_2$ by an angle $\delta \gamma$ to obtain the new vector $\mathbf{v}_1$ by using the base vector of the tangent space.
Based on the Rodrigues rotation formula, the updated direction can be denoted as
\begin{equation}
    \mathbf{v}_1 = \lambda \cdot \big( \cos(\delta\gamma)\mathbf{u}_1 + \sin(\delta\gamma)\mathbf{u}_3 \big),
    \label{eq:local_comp}
\end{equation}
which rotates $\mathbf{u}_1$ within the plane orthogonal to $\mathbf{v}_2$ by angle $\delta\gamma$. $\lambda \in \mathbb{R}_{+}$ is a scalar magnitude.

\begin{figure}
\includegraphics[width=\linewidth]{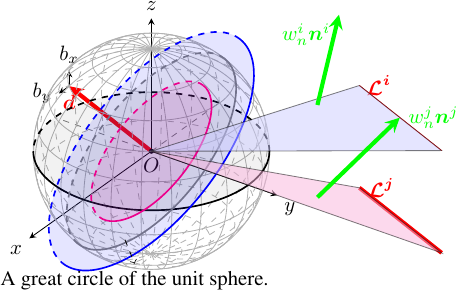}
\caption{Illustration of the 
parametrization for two parallel line landmarks $\mathbf{\mathcal{L}}^i$ and $\mathbf{\mathcal{L}}^j$. The vanishing direction vector $\mathbf{u}_2$ and normalized normal vector $\mathbf{u}_1$ are optimized on the tangent spaces $T_{\mathbf{u}_2}\mathcal{S}^2$ of the sphere and $T_{\mathbf{u}_1}\mathcal{S}^{1}$ of the circle, respectively. }
\label{fig:paral_line}
\end{figure}

\subsection{Unified Representation for Parallel Line Sets}~\label{sec:para-lines-repre}

As illustrated in Figure~\ref{fig:paral_line}, our method naturally extends to a set of parallel lines by building upon the representation of a single 3D line.
For a group of $k$ parallel lines denoted as $\mathbb{S} = \left[ \mathcal{L}^w_0, \mathcal{L}^w_1, \dots, \mathcal{L}^w_k \right]$, the shared vanishing direction is represented by a single unit vector $\mathbf{u}_2$. Each line $\mathcal{L}^w_i$ in the group is characterized by a local component $\mathbf{u}_1^i$ and a scale $\omega^i$. This results in a compact representation of the set as:
\begin{equation}
[\mathbf{u}_2, \; \omega^0\mathbf{u}_1^0, \; \omega^1\mathbf{u}_1^1, \; \dots, \; \omega^k\mathbf{u}_1^k],
\end{equation}
where all $\omega^{i}\mathbf{u}_1^i$ lying on the circle ${_{\omega^{i}}}\mathcal{S}^1$ perpendicular to $\mathbf{u}_2$. This leads to a minimal representation of the entire line group with $2 + 2k$ degrees of freedom: 2 DoF for the shared global direction $\mathbf{u}_2$, and 2 DoF per line for its local component.

As shown in Figure~\ref{fig:paral_line}, $\mathbf{u}_2$ lies on $\mathcal{S}^2$ and each $\mathbf{u}_1^i$ lies on the associated circle orthogonal to $\mathbf{u}_2$ by the minimal representation on the unit sphere.
Consequently, the minimal parameterization of the parallel line group follows the same format as the single-line case and can be written as:
\begin{equation}
    [\delta\bm{\theta}, \; \delta\gamma^0, \; \dots, \; \delta\gamma^k, \; \lambda^0, \; \dots, \; \lambda^k],
\end{equation}
where $\delta\bm{\theta}$ encodes the global direction $\mathbf{u}_2$, $\delta\gamma^i$ represents the angular perturbation of $\mathbf{u}_1^i$ on the circle, and $\lambda^i$ denotes the perturbation corresponding distance scale. 
Since parallel lines $\bm{\mathcal{L}}^i$ and $\bm{\mathcal{L}}^j$ share the same direction vector $\mathbf{u}_2$, the global vector $\mathbf{u}_2$ can thus be optimized via Equation~\ref{eq:delta_m} and \ref{eq:update_v2}.
The associated local vectors $\omega^i\bm{u}^i$ and $\omega^j\bm{u}^j$ lie on the same plane defined by this tangent plane, therefore these local components can be optimized separately. 

This unified formulation demonstrates the versatility of our approach that enables seamless extension from individual line landmarks to structurally consistent representations of parallel line groups within a common manifold-based optimization framework.


\section{Optimization with Points and Lines}

\subsection{Graph Construction}
The vertices in the point-line factor graphs $\mathcal{G}$ include camera poses $\mathcal{V}_{pose}$, point landmarks $\mathcal{V}_{p}$, line landmarks $\mathcal{V}_l$ and parallel-line sets $\mathcal{V}_{para}$. Specifically, camera pose $\mathbf{T}_{w,c_i} = \left[\begin{array}{cc}
    \mathbf{R}_{w,c_i} & \mathbf{t}_{w,c_i}  \\
     \mathbf{0} & 1 
\end{array}\right]$, where $\mathbf{T}_{w,c_i} \in SE(3)$, $\mathbf{R}_{w,c_i}\in SO(3)$, and $\mathbf{t}_{w,c_i} \in \mathbb{R}^3$. Points used in the optimization module is parametrized as $\mathbf{P}^k_w = \left[\begin{array}{ccc}
    x^k & y^k & z^k \\
\end{array}\right]^T$, and line landmarks are represented in minimal parameterization forms.

\begin{figure}
\centering
\begin{tabular}{cc}
    \resizebox{0.18\textwidth}{!}{%
    \begin{tikzpicture}[>=stealth',shorten >=2pt,node distance=2.0cm,  semithick]
\tikzstyle{posevertex}=[fill=black!25,minimum size=17pt,inner sep=0pt]
\tikzstyle{linevertex}=[fill=yellow!25,minimum size=17pt,inner sep=0pt]
\tikzstyle{parlivertex}=[fill=blue!25,minimum size=19pt,inner sep=0pt]
\tikzset{LineEdgeStyle/.style= {thick}}
\tikzset{ParaLiEdgeStyle/.style= {thick}}
\tikzset{LineLabelStyle/.style ={draw, fill= yellow, text= red}}
\tikzset{ParaLineLabelStyle/.style ={draw, fill= yellow, text= red}}
\foreach \name/\x in {1/1, 2/2, 3/3, 4/4, 5/5}
\node[posevertex] (pose-\name) at (\x,3) {$\name$};
\foreach \name/\x in {1/1.5, 2/2.5, 3/3.5, 4/4.5}
\node[linevertex] (line-\name) at (\x,0) {$\name$};
\draw[LineEdgeStyle](pose-1) to node[LineLabelStyle]{} (line-1);
\draw[LineEdgeStyle](pose-2) to node[LineLabelStyle]{} (line-1);
\draw[LineEdgeStyle](pose-3) to node[LineLabelStyle]{} (line-1);
\draw[LineEdgeStyle](pose-2) to node[LineLabelStyle]{} (line-3);
\draw[LineEdgeStyle](pose-4) to node[LineLabelStyle]{} (line-3);
\draw[LineEdgeStyle](pose-4) to node[LineLabelStyle]{} (line-2);
\draw[LineEdgeStyle](pose-5) to node[LineLabelStyle]{} (line-4);
\end{tikzpicture}}
&
  \resizebox{0.20\textwidth}{!}{%
    \begin{tikzpicture}[>=stealth',shorten >=2pt,node distance=2.0cm,
                    semithick]
\tikzstyle{posevertex}=[fill=black!25,minimum size=17pt,inner sep=0pt]
\tikzstyle{linevertex}=[fill=green!25,minimum size=17pt,inner sep=0pt]
\tikzstyle{paralinevertex}=[fill=red!25,minimum size=17pt,inner sep=0pt]
\tikzset{LineEdgeStyle/.style= {thick}}
\tikzset{ParaLiEdgeStyle/.style= {thick}}
\tikzset{LineLabelStyle/.style ={draw, fill= brown, text= red}}
\tikzset{ParaLineLabelStyle/.style ={draw, fill= pink, text= red}}
\foreach \name/\x in {1/1, 2/2, 3/3, 4/4, 5/5}
\node[posevertex] (pose-\name) at (\x,3) {$\name$};
\foreach \name/\x in {1/2, 2/3, 3/4, 4/5}
\node[linevertex] (line-\name) at (\x,0) {$\name$};
\node[paralinevertex] (para_0) at (0.5,0) {$0$};
\draw[LineEdgeStyle](pose-1) to node[LineLabelStyle]{} (line-1);
\draw[LineEdgeStyle](pose-2) to node[LineLabelStyle]{} (line-1);
\draw[LineEdgeStyle](pose-3) to node[LineLabelStyle]{} (line-1);
\draw[LineEdgeStyle](pose-2) to node[LineLabelStyle]{} (line-3);
\draw[LineEdgeStyle](pose-4) to node[LineLabelStyle]{} (line-3);
\draw[LineEdgeStyle](pose-4) to node[LineLabelStyle]{} (line-2);
\draw[LineEdgeStyle](pose-5) to node[LineLabelStyle]{} (line-4);
\path [bend left] (para_0) edge (line-1);
\path [bend left]   (para_0) edge (line-2);
\path [bend left] (para_0) edge (line-3);
\path [bend left]   (para_0) edge (line-4);
\end{tikzpicture}}
\end{tabular}
\\
\resizebox{0.45\linewidth}{!}{%
\centering
    \begin{tikzpicture}
    [>=stealth',shorten >=2pt,node distance=2.0cm, semithick]
	\tikzstyle{posevertex}=[fill=black!25,minimum size=17pt,inner sep=0pt]
	\tikzstyle{linevertex}=[fill=yellow!25,minimum size=17pt,inner sep=0pt]	
	\tikzstyle{paral1vertex}=[fill=green!25,minimum size=17pt,inner sep=0pt]
	\tikzstyle{paralinevertex}=[fill=red!25,minimum size=17pt,inner sep=0pt]
	\tikzset{LineEdgeStyle/.style= {thick}}
	\tikzset{ParaLiEdgeStyle/.style= {thick}}
	\tikzset{ParaLineSharedLabelStyle/.style ={draw, fill= pink, text= black}}
	\tikzset{RepreojLabelStyle/.style ={draw, fill= yellow, text= black}}
	\tikzset{ParaLineUniqueLabelStyle/.style ={draw, fill= brown, text= black}}
	\node[posevertex] (pose_i) at (-1.0,2.0){} ;
	\node[linevertex] (line_i) at (3.0,2.0){}  ;
	\draw[] (-0.7,2.0) node[anchor=west] {\huge Camera Pose};
	\draw[] (3.3,2.0) node[anchor=west] {\huge Parallel Lines};
	\node[paral1vertex] (line_j) at (-1.0,1.2){} ;
	\node[paralinevertex] (para_0) at (3,1.2) {};
	\draw[] (-0.7,1.2) node[anchor=west] {\huge Local Info.};
	\draw[] (3.3,1.2) node[anchor=west] {\huge Global Info.};
	\end{tikzpicture}} 
\resizebox{0.4\linewidth}{!}{%
\centering
    \begin{tikzpicture}[>=stealth',shorten >=2pt,node distance=2.0cm,
                    semithick]
\tikzstyle{posevertex}=[fill=black!25,minimum size=17pt,inner sep=0pt]
\tikzstyle{linevertex}=[fill=yellow!25,minimum size=17pt,inner sep=0pt]
\tikzstyle{paral1vertex}=[fill=green!25,minimum size=17pt,inner sep=0pt]
\tikzstyle{paralinevertex}=[fill=red!25,minimum size=17pt,inner sep=0pt]
\tikzset{LineEdgeStyle/.style= {thick}}
\tikzset{ParaLiEdgeStyle/.style= {thick}}

\tikzset{RepreojLabelStyle/.style ={draw, fill= yellow, text= black}}
\tikzset{ParaLineUniqueLabelStyle/.style ={draw, fill= brown, text= black}}
\draw[RepreojLabelStyle](-0.2,1.5) to node[RepreojLabelStyle]{} (0.5,1.5) node[anchor=west]{\large Reprojection Factor};
\draw[ParaLineUniqueLabelStyle](-0.2,1.8) to node[ParaLineUniqueLabelStyle]{} (0.5,1.8) node[anchor=west]{\large Reprojection Factor};
\end{tikzpicture}} \\
\caption{Factor graph representations for different line-based structures. Left: conventional line re-projection factors. Right: the proposed parallel line representation, explicitly separating global and local components with re-projection factors.}
\label{fig:factor_graphs}
\end{figure}
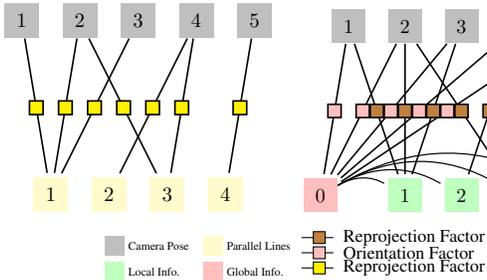

\subsection{Factors and Constraints}
Co-visibility connections~\cite{mur2015orb,mur2017orb,campos2021orb} are built when two images detect the same landmark, such as a point or a line on the map. 
In this section, we define the co-visibility factors based on the re-projection models of point and line features.

\textbf{Co-visibility factors from points.}
Based on the point feature measurement model, the measurement of the $k^{th}$ global point landmark $\mathbf{P}_w^{k}$ at frame $c_j$ is represented as $\mathbf{\bar{p}}_{k}^{j}$ in the normalized coordinate, and the re-projection factor of a point feature is defined as $\mathbf{r}_{\mathbf{p}}(\mathbf{\bar{p}}_i^{j}, \mathbf{P}_w^k, \mathbf{T}_{w,c_j})$.

\textbf{Co-visibility factors from lines.} \label{traditional_line}
Traditionally, the mapline $\mathcal{L}_w^i$ in the world coordinate is transferred to the camera coordinates, and then re-projected on the image plane $\mathbf{l}^j_k$ of viewpoint $c_j$. 
The error between the re-projected line $\mathbf{l}^j_k$ and the two endpoints $\mathbf{\bar{p}}^j_{k,s}$ and $\mathbf{\bar{p}}^j_{k,e}$ of the extracted 2D line can be written as:
\begin{equation}
    \mathbf{r}_{\mathbf{l}}( \mathbf{\bar{p}}^j_{k,s},\mathbf{\bar{p}}^j_{k,e}, \mathcal{L}^k_w, \mathbf{T}_{w,c_j} )
    = \left[\begin{array}{c}
         \operatorname{dis}(\mathbf{\bar{p}}^j_{k,s}, \mathbf{l}^j_k )  \\
         \operatorname{dis}(\mathbf{\bar{p}}^j_{k,e}, \mathbf{l}^j_k )
    \end{array}\right],
\end{equation}
where $\operatorname{dis}(\cdot)$ gives the distance between a point and a line.

\textbf{Constraints between parallel lines.}
To enforce structural consistency, we incorporate additional constraints among groups of parallel lines. Specifically, for each mapline $\mathcal{L}_w^i$ in a set of $N$ parallel lines $\{\mathcal{L}_w^i\}_{i=1}^N$, its direction should be parallel to other lines in the set. The parallelism constraint is enforced by minimizing the angular deviation between the direction of each line $\mathbf{u}_2^i$. The residual is defined as:
\begin{equation}
    \mathbf{r}_{\parallel}(\mathcal{L}_i^w, \{\mathcal{L}_j^w\}_{j=1, j\neq i}^N ) =\frac{1}{N-1} \sum_{j=1, j\neq i}^{N} \left(1 - \mathbf{u}_2^{i\top} \mathbf{u}_2^j \right).
\end{equation}
Here, the residual $\mathbf{r}_{\parallel}$ encourages all lines in the group to remain parallel to each other during optimization.

\begin{table*}
\centering
\resizebox{\linewidth}{!}{
\begin{tabular}{l|cc|cccccccc|cccc}
\toprule
\multirow{3}{*}{\textbf{Sequence}} &\multicolumn{2}{c|}{ \textbf{Initial} } & \multicolumn{8}{c|}{ \textbf{Optimization using Independent Primitives} }  & \multicolumn{4}{c}{ \textbf{Optimization using Structure Constraints } } \\
& \multicolumn{2}{c|}{ Factor Graph}
& \multicolumn{2}{c}{ Point  }
& \multicolumn{2}{c}{ OrthLine }
& \multicolumn{2}{c}{ ImplicitLine }
& \multicolumn{2}{c|}{RiemanLine}
& \multicolumn{2}{c}{OrthLine\_Constr}
& \multicolumn{2}{c}{StructRiemanLine}\\ 
&RMSE &Med.   &RMSE  &Med.  &RMSE  &Med.  &RMSE  &Med.  &RMSE  &Med.  &RMSE  &Med. &RMSE  &Med.  \\ \hline
livingroom0 &12.13 &1.95 & 1.14 &0.69 &0.81 &0.38 &0.96 &0.40 &\textbf{0.81} &0.38 &\textbf{0.81} &0.38 &\textbf{0.81} &\textbf{0.37}\\
livingroom1 &2.15 &1.57   &1.67 &1.24  
            &4.13 &4.22    &0.92 &0.81
            &1.19 &0.71   &3.58 &3.60 &\textbf{0.90} &\textbf{0.55} \\              
livingroom2 &12.95 &6.95  &1.61 &0.80   
            &0.79 &0.71  &1.18 &0.90            
            &0.78 &0.71 &0.79 &0.72 &\textbf{0.75} &\textbf{0.68}\\            
livingroom3 &18.26 & 13.93 &11.64 &11.43 &6.70 &6.33 &6.67 &6.76 &6.71 &6.34 &6.71 &6.34 &\textbf{6.65} &\textbf{6.33}\\    

office0     &1.73 &0.69 &0.70 &0.34      
           &\textbf{0.38} &\textbf{0.26} &0.47 &0.37   
           &\textbf{0.38} &\textbf{0.26}  &\textbf{0.38} &\textbf{0.26} &\textbf{0.38} &\textbf{0.26}  \\            
office1    &11.93 &5.48 &7.68 &5.22  
           &1.81 &0.89 &9.78 &7.81 &1.89 &0.87 &1.81 &0.87 &\textbf{1.76} &\textbf{0.73}  \\ 
       
office2 &2.95 &1.88  &0.93 &0.55 &\textbf{0.75} &0.56 &\textbf{0.75} &\textbf{0.53} &\textbf{0.75} &0.56 &0.76 &0.56 &\textbf{0.75} &0.56\\
office3 &7.56 &2.47  & 1.22 &0.66 &0.78 &0.49 &1.86 &0.77 &0.79 &0.51 &0.85 &0.53 &\textbf{0.77} &\textbf{0.46} \\
\bottomrule  
\end{tabular}}
\caption{Comparison of translation (APE) RMSE and median errors (cm) on the ICL-NUIM~\cite{handa2014benchmark} benchmark dataset. The best results are highlighted in \textbf{bold}.}
\label{tab:ape_icl}
\end{table*}

\section{Experiments}~\label{sec:experiment}

We evaluate the proposed RiemanLine and StructRiemanLine parameterizations on both public benchmarks and simulated environments. Our goals are to assess: (i) pose estimation accuracy, (ii) landmark reconstruction quality, and (iii) the benefits of incorporating structural constraints within a unified Riemannian manifold representation.

\textbf{Baselines.} We compare against standard parameterizations widely used in SLAM and VO: \textbf{Euclidean XYZ}~\cite{mur2015orb}, \textbf{Orthonormal}~\cite{bartoli2005structure}, and \textbf{ImplicitLine}~\cite{Zhao_Huang_Yan_Dissanayake_2015}. The Orthonormal form serves as the minimal representation for Closest Point~\cite{8793507} and Quaternion~\cite{kottas2013efficient} methods. All baselines are fed with the same co-visibility factor graph in each sequence.

\textbf{Datasets.} We use the ICL-NUIM benchmark~\cite{handa2014benchmark} (eight indoor RGB-D sequences) and four TartanAir~\cite{wang2020tartanair} sequences featuring photorealistic synthetic environments. Additionally, we evaluate on three challenging simulation sequences (corridor, box, sphere) generated using Open-Structure~\cite{li2023open}.

\textbf{Metrics.} We report Absolute Trajectory Error (ATE) for camera localization, and angular errors (direction and normal) for line reconstruction. All results are averaged per sequence. Computations were performed on an Intel NUC Mini PC with Core i7-8700 CPU.

\subsection{Evaluation on ICL-NUIM}
Table~\ref{tab:ape_icl} summarizes ATE RMSE and MEDIAN errors. The initial factor graphs show significant drift (e.g., RMSE $12.13\,\text{cm}$ on \textit{livingroom0}). 
For brevity, we refer to \textbf{Point\_StructRiemanLine} as \textbf{StructRiemanLine} in tables.
The proposed \textbf{Point\_StructRiemanLine} achieves the best overall accuracy across all eight sequences. By explicitly encoding parallel-line constraints into the minimal Riemannian representation, it reduces errors without introducing additional parameters. For example, \textit{livingroom2} achieves RMSE $0.75\,\text{cm}$ (MEDIAN $0.68\,\text{cm}$), and \textit{office2} maintains RMSE $0.75\,\text{cm}$ while preserving robust convergence behavior.

\begin{table}[!t]
\centering
\setlength{\tabcolsep}{1.1pt}
\resizebox{\linewidth}{!}{
\begin{tabular}{l|cc|cc|cc|cc}
\hline
\multirow{2}{*}{Sequence} & \multicolumn{2}{c|}{Carwelding} & \multicolumn{2}{c|}{Hospital} & \multicolumn{2}{c|}{Office} & \multicolumn{2}{c}{Jpn. Alley} \\
 & Trans. & Rot. & Trans. & Rot. & Trans. & Rot. & Trans. & Rot. \\
\hline
Initial & 14.74 & 10.89 & 33.78 & 26.83 & 57.00 & 47.78 & 10.92 & 5.23 \\ 
OrthLine & 4.47 & 0.16 & 5.32 & 4.83 & 43.25 & 29.23 & 3.75 & 2.80 \\ 
OrthLine\_Const & 4.56 & 0.16 & 5.49 & 4.97 & 50.04 & 39.50 & 3.68 & 2.72 \\ 
RiemanLine & 4.46 & 0.16 & 7.29 & 6.60 & 40.95 & 26.70 & 3.85 & 2.92 \\ 
StructRiemanLine & \textbf{4.08} & \textbf{0.15} & \textbf{2.91} & \textbf{2.04} & \textbf{15.60} & \textbf{12.08} & \textbf{3.65} & \textbf{2.68} \\
\hline
\end{tabular}}
\caption{Comparison of APE RMSE (translation (cm) and rotation (degree) on the TartanAir dataset.}
\label{tab:tartanair_trans_rot}
\end{table}









\begin{table}[!t]
\centering
\small
\setlength{\tabcolsep}{1.1pt}
\renewcommand{\arraystretch}{1}
\resizebox{\linewidth}{!}{
\begin{tabular}{l|cc|cc|cc|cc}
\hline
\multirow{2}{*}{Method}
& \multicolumn{2}{c|}{Carweld.}
& \multicolumn{2}{c|}{Hospital}
& \multicolumn{2}{c|}{Office}
& \multicolumn{2}{c}{Jpn. Alley} \\
& Dir. & Norm.
& Dir. & Norm.
& Dir. & Norm.
& Dir. & Norm. \\
\hline
OrthLine          & 1.52 & 1.42 & 5.16 & 4.11 & 3.89 & 4.49 & 0.88 & 0.86 \\
OrthLine\_Constr  & 0.92 & 0.93 & 5.05 & 2.54 & 2.04 & 2.28 & 0.46 & 0.42 \\
RiemanLine        & 1.55 & 1.43 & 5.38 & 4.68 & 3.89 & 4.48 & 0.88 & 0.73 \\
StructRiemanLine  & \textbf{0.82} & \textbf{0.83} & \textbf{1.60} & \textbf{0.91} & \textbf{0.04} & \textbf{0.51} & \textbf{0.07} & \textbf{0.15} \\
\hline
\end{tabular}}
\caption{Comparison of line reconstruction performance on TartanAir. Median errors (degrees) are reported for direction (Dir.) and normal (Norm.). }
\label{tab:line_reconstruction}
\end{table}


\subsection{Evaluation on TartanAir}

Table~\ref{tab:tartanair_trans_rot} reports ATE translation and rotation errors on the TartanAir benchmark. 
The proposed \textbf{StructRiemanLine} achieves the most significant improvement in structurally rich indoor sequences. 
On \textit{Hospital}, the translation RMSE drops from $33.78$ cm (Initial) and $7.29$ cm (\textbf{RiemanLine}) 
to only $2.91$ cm, while the rotation RMSE decreases from $26.83^\circ$ to $2.04^\circ$. 
Similarly, on \textit{Office}, the translation error is reduced by over \textbf{40\%} compared to Point\_OrthLine and Point\_RiemanLine, demonstrating that the proposed parallel-line manifold representation provides strong orientation priors that are especially beneficial in man-made environments with dominant structural regularities. 
These results validate that explicitly encoding shared vanishing directions within the factor graph not only improves translational accuracy but also significantly enhances rotational consistency, a critical factor for large indoor scene reconstruction.

\begin{figure}
    \centering
    \includegraphics[width=\linewidth]{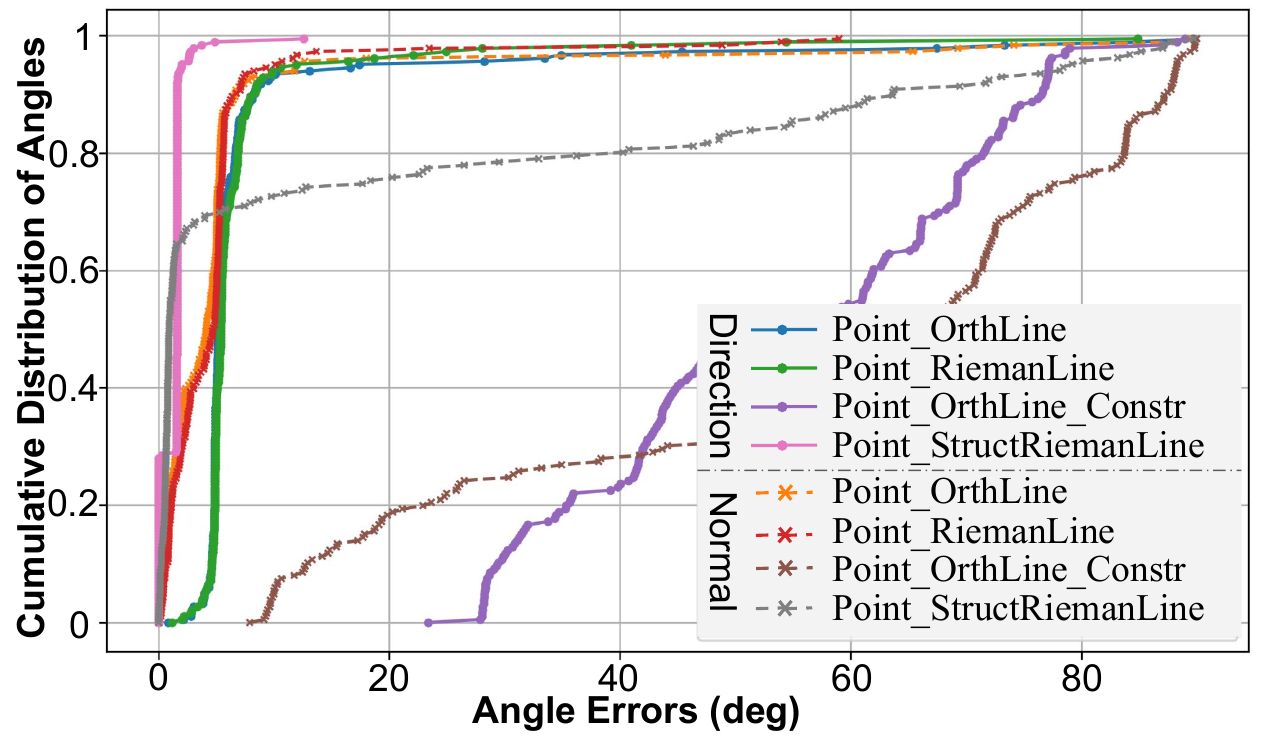}
    \caption{Line reconstruction errors of different methods in the \textit{Hospital} sequence.}
    \label{fig:line_reconstruction}
\end{figure}

\textbf{Line reconstruction.} Table~\ref{tab:line_reconstruction} reports median angular errors. StructRiemanLine achieves the lowest direction and normal errors across all TartanAir sequences. In \textit{Hospital}, the direction error is reduced to $1.60^\circ$ and the normal error to $0.91^\circ$, validating that structural encoding improves both pose estimation and landmark quality. Figure~\ref{fig:line_reconstruction} shows the cumulative distribution of angular errors for line direction and normal estimation on the \textit{Hospital} sequence. The proposed StructRiemanLine achieves the steepest rise and highest saturation, indicating that over $90\%$ of reconstructed lines fall within $2^\circ$ of the ground truth for both direction and normal vectors. 
Compared to Point\_Orthonormal and Point\_RiemanLine, which exhibit heavier tails beyond $5^\circ$, StructRiemanLine demonstrates significantly reduced variance. This confirms that encoding parallelism within the Riemannian manifold not only lowers median error (Table~\ref{tab:line_reconstruction}) but also improves the consistency of line reconstruction.

\begin{figure*}[!t]
    \centering
    \includegraphics[width=\linewidth, trim={0 10 20 5},clip]{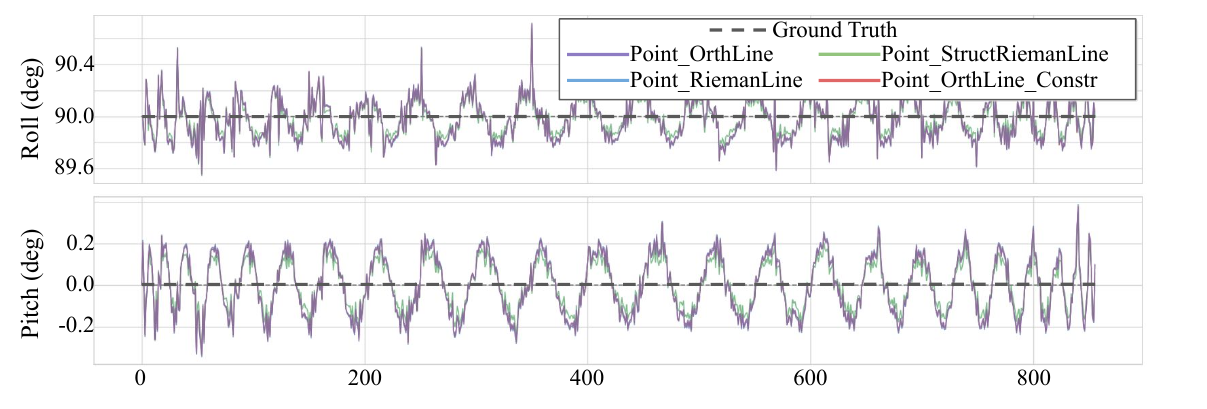}
    \caption{Pitch and roll angle errors for the sphere sequence in the simulation dataset. The plots compare the estimated orientations of different methods against the ground truth.}
    \label{fig:simulation_sphere}
\end{figure*}

\subsection{Simulations}

\begin{figure}
    \centering
    \includegraphics[width=\linewidth]{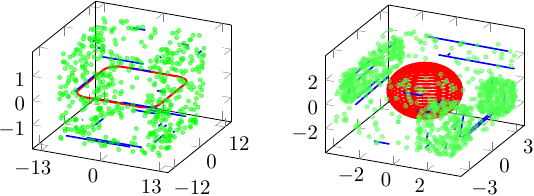}
    \caption{Visualization of the simulation scenarios: \textit{corridor} and \textit{sphere}. Points, lines, and trajectories are shown in green, blue, and red, respectively.}
    \label{fig:simulation_visual}
\end{figure}

Figure~\ref{fig:simulation_visual} illustrates the synthetic environments that are \textit{corridor} and \textit{sphere}. In the corridor scenario, a rectangular arrangement of green points forms a corridor-like structure with blue structural lines embedded within, while a red trajectory traces a loop along the corridor’s center. The sphere scenario depicts a roughly spherical distribution of green points surrounded by blue lines, where the red trajectory forms a dense circular pattern around the sphere’s surface.

Table~\ref{tab:simulation} reports ATE RMSE. While both RiemanLine and Orthonormal perform similarly on simple structures, StructRiemanLine consistently achieves the lowest translation and rotation errors. In the \textit{sphere} scenario, it reduces translation RMSE to $1.16\,\text{cm}$ and maintains minimal orientation drift, confirming the benefits of enforcing parallelism constraints in structurally symmetric environments.

Figure~\ref{fig:simulation_sphere} further compares pitch and roll estimation errors in the \textit{sphere} simulation. Point\_Orthonormal and Point\_RiemanLine exhibit small but noticeable oscillations, while StructRiemanLine maintains the closest alignment with ground truth across the trajectory. 



\begin{table}[!t]
\centering
\setlength{\tabcolsep}{1.1pt}
\resizebox{\linewidth}{!}{
\begin{tabular}{l|cc|cc|cc}
\hline
\multirow{2}{*}{Method} & \multicolumn{2}{c|}{corridor} & \multicolumn{2}{c|}{box} & \multicolumn{2}{c}{sphere} \\
 & Trans. & Rot.  & Trans. & Rot. & Trans. & Rot.  \\
\hline
Initial  & 10.57 & 1.54   & 9.67 &0.85  &25.99 &5.49\\ 
OrthLine & 4.24 &0.32  &2.09 &\textbf{0.23} &1.19 &\textbf{0.19} \\ 
OrthLine\_Constr & 4.29 & 0.32 &2.09 &\textbf{0.23} &1.19 &\textbf{0.19}\\ 
RiemanLine & 4.24 & 0.32  &2.09 &\textbf{0.23}  &1.20 &\textbf{0.19} \\ 
StructRiemanLine& \textbf{4.02} &\textbf{0.31}  &\textbf{2.08} &\textbf{0.23} &\textbf{1.16} & \textbf{0.19}\\
\hline
\end{tabular}}
\caption{Comparison of translation (APE RMSE (cm)) and rotation (APE RMSE (degree)) on the simulation dataset.}
\label{tab:simulation}
\end{table}

\textbf{Runtime and Complexity Analysis.}
Table~\ref{tab:complex_time} reports the parameter complexity and solver runtime 
on the \textit{box} simulation sequence, which contains $447$ camera poses, $130$ point landmarks, 
$20$ line landmarks, and $2$ parallel-line groups. 
Landmarks observed by fewer than three cameras are pruned prior to optimization 
to ensure a consistent and well-constrained factor graph. Starting from the \textit{Point-only} baseline, the inclusion of $20$ line landmarks in 
{Point\_OrthLine\_Constr} introduces $20$ additional parameter blocks and 
\textbf{80} new parameters due to the conventional $4n$ minimal parameterization. 
In contrast, the proposed \textbf{StructRiemanLine} requires only \textbf{44} parameters 
to represent the same $20$ lines, as $19$ of them are encoded within two parallel-line groups under the compact $2n+2$ formulation.

In terms of computational efficiency, the total optimization time drops from 
\textbf{97.62~s} to \textbf{49.62~s}, yielding an approximate 49\% overall speed-up. This improvement is consistent with the reduction in parameter blocks 
($597 \rightarrow 579$) and state dimensionality 
($3152 \rightarrow 3116$), which together produce a sparser and better-conditioned Hessian.

\begin{table}
    \centering
    \setlength{\tabcolsep}{1.1pt}
    \resizebox{\linewidth}{!}{
    \begin{tabular}{l|ccccc}
    \toprule
       Method & Para. Blks. & Eff. Params. & Resi. Blks. & Resi. & Time (s) \\ \hline
       \textit{Point} & \textit{577}   &\textit{3072}  & \textit{36010} &\textit{37010}  &\textit{53.90}\\  \hdashline
       OrthLine\_Constr &597 & 3152& 41773 &83452 &97.62 \\ 
       StructRiemanLine &579 &3116 &41679 & 83358 &49.62 \\
       \bottomrule
    \end{tabular}}
    \caption{Comparison of parameter complexity and total solver runtime between 
the conventional Orthonormal representation with additional parallelism constraints and the proposed minimal parametrization on the \textit{box} sequence}
    \label{tab:complex_time}
\end{table}

\section{Conclusion and Future Works}
We have presented a novel minimal representation for 3D line landmarks. 
Unlike conventional approaches, the proposed parameterization not only encodes individual lines with minimal degrees of freedom, but also naturally accommodates structural regularities such as sets of parallel lines. 
Building upon this representation, we introduced a joint factor graph framework that integrates both co-visibility and structural factors, enabling more accurate and efficient optimization of camera poses and landmarks compared to traditional point–line co-visibility graphs.

Looking ahead, we envision incorporating the proposed parameterization into full SLAM and visual odometry pipelines to achieve more robust tracking and high-fidelity reconstruction in large-scale environments. 

\section*{Acknowledgments}
We would like to thank Xin Li for insightful discussions and constructive feedback. 
This research was supported by the Tier 2 Grant (MOE-T2EP20124-0015) from the Singapore Ministry of Education.

\bibliography{aaai2026}

\end{document}